\documentclass [final] {IEEEtran}

\usepackage{multirow}

\usepackage{algorithm}
\usepackage{algorithmic}

\makeatother
\usepackage[utf8]{inputenc}
\usepackage[table]{xcolor}
\usepackage{tabularx,booktabs}

\newcolumntype{C}{>{\centering\arraybackslash}X} 
\usepackage{graphics} 
\usepackage{epsfig} 
\usepackage{amsmath} 
\usepackage{amssymb}  

\usepackage{wrapfig}
\usepackage{times}
\usepackage{cite}
\usepackage{url}
\usepackage{subfigure}
\graphicspath{{sHEMO_Figures/}}
\usepackage{pifont}
\usepackage{xcolor}
\usepackage{rotating}
\usepackage{rotfloat}
\usepackage{algorithm}
\usepackage{algorithmic}
\usepackage{gensymb}
\usepackage{textcomp}
\usepackage{placeins}
\usepackage{balance}
\usepackage{subfigure}
\usepackage{booktabs}
\usepackage{tabularx}
\usepackage{array}
\usepackage{pbox}
\usepackage{longtable}
\usepackage{booktabs}
\usepackage[font=small,skip=0pt]{caption}
\usepackage{url}

\begin{document}


\title{Advancing Smart Malnutrition Monitoring: A Multi-Modal Learning Approach for Vital Health Parameter Estimation}

\author{Ashish Marisetty, Prathistith Raj M, Praneeth Nemani, Venkanna Udutalapally, Debanjan Das
}

\maketitle

\begin{abstract}
Malnutrition poses a significant threat to global health, resulting from an inadequate intake of essential nutrients that adversely impacts vital organs and overall bodily functioning. Periodic examinations and mass screenings, incorporating both conventional and non-invasive techniques, have been employed to combat this challenge. However, these approaches suffer from critical limitations, such as the need for additional equipment, lack of comprehensive feature representation, absence of suitable health indicators, and the unavailability of smartphone implementations for precise estimations of Body Fat Percentage (BFP), Basal Metabolic Rate (BMR), and Body Mass Index (BMI) to enable efficient smart-malnutrition monitoring. To address these constraints, this study presents a groundbreaking, scalable, and robust smart malnutrition-monitoring system that leverages a single full-body image of an individual to estimate height, weight, and other crucial health parameters within a multi-modal learning framework. Our proposed methodology involves the reconstruction of a highly precise 3D point cloud, from which 512-dimensional feature embeddings are extracted using a headless-3D classification network. Concurrently, facial and body embeddings are also extracted, and through the application of learnable parameters, these features are then utilized to estimate weight accurately. Furthermore, essential health metrics, including BMR, BFP, and BMI, are computed to conduct a comprehensive analysis of the subject's health, subsequently facilitating the provision of personalized nutrition plans. While being robust to a wide range of lighting conditions across multiple devices, our model achieves a low Mean Absolute Error (MAE) of $\pm$ 4.7 cm and $\pm$ 5.3 kg in estimating height and weight.

\end{abstract}

\begin{IEEEkeywords}
Multi-modal Learning, 3D Reconstruction, Feature Fusion, Height and Weight estimation, Smart Healthcare, Non-invasive.
\end{IEEEkeywords}


\maketitle

\section{Introduction}
\textbf{Malnutrition} is an ailment caused by consuming food that lacks an adequate quantity of essential nutrients. It is most commonly used in reference to undernutrition, \cite{10.1007/978-981-10-9059-2_10} which occurs when a person does not receive sufficient calories, proteins, or micronutrients. A scarcity of a quality diet most commonly causes undernourishment or undernutrition. According to a WHO survey, there are 178 million malnourished children globally, with 20 million suffering from severe malnutrition, contributing to 3.5 to 5 million deaths in children under five each year. On a global scale, undernutrition is responsible for 45\% of all casualties in children under five and is widespread in developing nations, especially among women and children. Malnutrition also poses a range of severe health problems that include anemia, diarrhea, disorientation, weight loss, night blindness, anxiety, attention deficits, and other neuropsychologic disorders \cite{8998172}. In the aftermath of the COVID-19 outbreak, which caused significant concerns and stress regarding public health \cite{9565356}, the traditional approach of measuring height and weight in public health centers has been impacted. During the pandemic, strict social distancing measures were put in place to minimize the spread of infection, making the conventional method of calculating essential health metrics through direct measurements undesirable.

\begin{figure}[htbp]
    \centering
    \includegraphics[width=\linewidth]{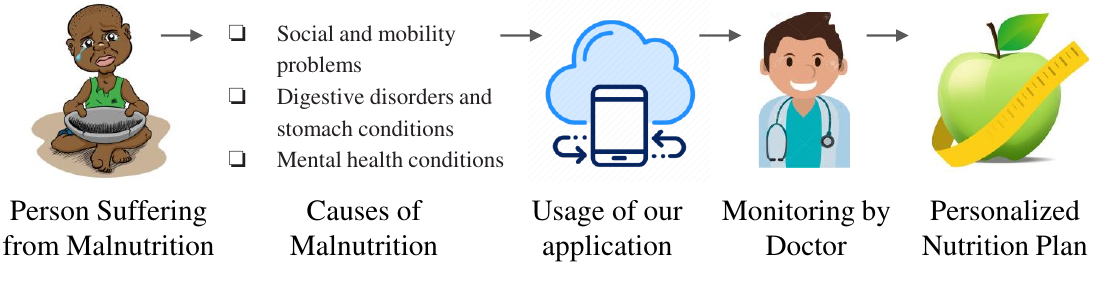}
    \caption{Conceptual Overview}
    \label{Conceptual Overview}
\end{figure}

In addition, pandemics like COVID-19, according to UNICEF, put malnourished children at an ever-increasing danger of mortality, as well as impaired growth, development, and learning for those who survive. Therefore, there is a dire need to identify important health indicators and monitor chronic stress \& uncontrolled or unmonitored food consumption integrated with data-driven approaches \cite{9184922}. A primary step in identifying or diagnosing malnutrition and the nutritional status of any person can be determined by computing their \textbf{Body Fat Percentage (BFP)}, \textbf{Basal Metabolic Rate (BMR)}, and \textbf{Body Mass Index (BMI)} and comparing it with standardized charts. It is more accurate to infer the risk of malnutrition and various medical conditions from these metrics since they represent the human body's functionality in a well-oriented manner. In this work, we intend to predict the height, weight and successively calculate the important health metrics as mentioned above from a sing-shot full-body image by incorporating a holistic representation of prominent features under the multi-modal learning paradigm. Fig. \ref{Conceptual Overview} illustrates a conceptual overview of the proposed method.

\begin{table*}[t]
 \caption{Comparison with existing literature works}
 \label{Related Research Works}

\begin{tabularx}{\textwidth}{@{}l*{11}{C}c@{}}
\toprule
Existing Technologies& 
Height Estimation& 
Weight Estimation& 
Holistic Feature Representation& 
Local 3D Features& 
Smartphone Application& 
Real-Time Testing& 
Other Health Metrics\\ 
\midrule
Alberink {\it et al.} \cite{alberink2008obtaining} 
&$\textcolor{green}{\checkmark}$ 
&$\textcolor{red}{\times}$ 
&$\textcolor{red}{\times}$ 
&$\textcolor{red}{\times}$ 
&$\textcolor{red}{\times}$
&$\textcolor{red}{\times}$ 
&$\textcolor{red}{\times}$ \\

Abdelkader {\it et al.} \cite{4813453} 
&$\textcolor{green}{\checkmark}$ 
&$\textcolor{red}{\times}$ 
&$\textcolor{green}{\checkmark}$
&$\textcolor{red}{\times}$ 
&$\textcolor{red}{\times}$ 
&$\textcolor{green}{\checkmark}$ 
&$\textcolor{red}{\times}$ \\

Dey {\it et al.}   \cite{10.1145/2660460.2660466} 
&$\textcolor{green}{\checkmark}$ 
&$\textcolor{red}{\times}$ 
&$\textcolor{red}{\times}$ 
&$\textcolor{red}{\times}$ 
&$\textcolor{red}{\times}$  
&$\textcolor{red}{\times}$ 
&$\textcolor{red}{\times}$ \\

Dantcheva \textit{et al.} \cite{8546159}
&$\textcolor{green}{\checkmark}$
&$\textcolor{green}{\checkmark}$
&$\textcolor{red}{\times}$ 
&$\textcolor{red}{\times}$ 
&$\textcolor{red}{\times}$
&$\textcolor{green}{\checkmark}$  
&$\textcolor{green}{\checkmark}$ \\

Gunel \textit{et al.} \cite{gunel2019face} 
&$\textcolor{green}{\checkmark}$ 
&$\textcolor{red}{\times}$ 
&$\textcolor{green}{\checkmark}$ 
&$\textcolor{red}{\times}$ 
&$\textcolor{red}{\times}$  
&$\textcolor{red}{\times}$ 
&$\textcolor{red}{\times}$ \\

Fukun {\it et al.} \cite{9156541} 
&$\textcolor{green}{\checkmark}$ 
&$\textcolor{red}{\times}$ 
&$\textcolor{green}{\checkmark}$ 
&$\textcolor{red}{\times}$ 
&$\textcolor{red}{\times}$
&$\textcolor{red}{\times}$ 
&$\textcolor{red}{\times}$ \\

Lee {\it et al.} \cite{lee2020human} 
&$\textcolor{green}{\checkmark}$ 
&$\textcolor{red}{\times}$ 
&$\textcolor{green}{\checkmark}$ 
&$\textcolor{red}{\times}$ 
&$\textcolor{red}{\times}$
&$\textcolor{red}{\times}$ 
&$\textcolor{red}{\times}$ \\

Velardo {\it et al.} \cite{5634540} 
&$\textcolor{red}{\times}$ 
&$\textcolor{green}{\checkmark}$
&$\textcolor{green}{\checkmark}$
&$\textcolor{red}{\times}$ 
&$\textcolor{red}{\times}$ 
&$\textcolor{red}{\times}$ 
&$\textcolor{red}{\times}$ \\

Nguyen \textit{et al.} \cite{nguyen2014seeing}
&$\textcolor{red}{\times}$ 
&$\textcolor{green}{\checkmark}$
&$\textcolor{green}{\checkmark}$
&$\textcolor{red}{\times}$ 
&$\textcolor{red}{\times}$ 
&$\textcolor{red}{\times}$ 
&$\textcolor{red}{\times}$ \\

Jiang {\it et al.} \cite{8666768}
&$\textcolor{red}{\times}$ 
&$\textcolor{green}{\checkmark}$ 
&$\textcolor{green}{\checkmark}$
&$\textcolor{red}{\times}$ 
&$\textcolor{red}{\times}$ 
&$\textcolor{green}{\checkmark}$  
&$\textcolor{red}{\times}$ \\

Jin \textit{et al.} \cite{jin2022estimating}
&$\textcolor{green}{\checkmark}$ 
&$\textcolor{green}{\checkmark}$ 
&$\textcolor{green}{\checkmark}$
&$\textcolor{green}{\checkmark}$ 
&$\textcolor{red}{\times}$ 
&$\textcolor{green}{\checkmark}$  
&$\textcolor{green}{\checkmark}$ \\

Altinigne \textit{et al.} \cite{9053363}
&$\textcolor{green}{\checkmark}$
&$\textcolor{green}{\checkmark}$
&$\textcolor{red}{\times}$ 
&$\textcolor{red}{\times}$ 
&$\textcolor{red}{\times}$
&$\textcolor{red}{\times}$  
&$\textcolor{red}{\times}$ \\ 

Thapar \textit{et al.} \cite{7906819}
&$\textcolor{red}{\times}$
&$\textcolor{red}{\times}$
&$\textcolor{red}{\times}$
&$\textcolor{red}{\times}$
&$\textcolor{red}{\times}$
&$\textcolor{red}{\times}$ 
&$\textcolor{red}{\times}$ \\ 

Child Growth Monitor \cite{Microsoft}
&$\textcolor{green}{\checkmark}$ 
&$\textcolor{green}{\checkmark}$
&$\textcolor{red}{\times}$
&$\textcolor{red}{\times}$ 
&$\textcolor{green}{\checkmark}$
&$\textcolor{green}{\checkmark}$ 
&$\textcolor{red}{\times}$ \\ 
\bottomrule

\textbf{{\it \bf auto}Nutri}
&$\textcolor{green}{\checkmark}$
&$\textcolor{green}{\checkmark}$
&$\textcolor{green}{\checkmark}$
&$\textcolor{green}{\checkmark}$
&$\textcolor{green}{\checkmark}$
&$\textcolor{green}{\checkmark}$ 
&$\textcolor{green}{\checkmark}$ \\ 
\bottomrule

\end{tabularx}

\end{table*}

In this paper, we propose a solution based on multi-feature fusion that includes 3D, facial, body, and metadata features integrated with a smartphone application prototype to estimate a human's height, weight, and other health parameters. The smartphone's camera serves as a sensor to capture a full-body image of a human, and the height is estimated by calculating the centimetre per pixel ratio using image processing techniques. Following that, the captured image is pre-processed by detecting, cropping, aligning the face and body, reconstructing \& samping a 3D person mesh object, and feature extraction in a multi-modal framework. To summarize, the key contributions of our work are:


\subsection{Contributions}

\begin{itemize}
  \item A holistic feature fusion of facial, body \& 3D embeddings, including the correlation between them, optimal feature combination and individual importance in estimating the weight is insightfully discussed.
  \item This paper is the first to incorporate the fine-grain local 3D representation in combination using 3D classification network backbones as feature extractors.
  \item To the best of our knowledge, this is the first time an IoMT framework has been used to develop an autonomous smart application for peripheral devices without any manual intervention.
  \item The trained model outperformed state-of-the-art methods for weight estimation on real-world data using a multi-modal architecture, achieving a 5.3 kg error.
\end{itemize}

\section{Related Research Overview}
With the COVID-19 pandemic behind us and a shift in the global landscape, including a rise in obesity and undernutrition in many countries, the need for a simple non-contact height and weight estimation technique remains as relevant as ever. Ongoing research is actively investigating and developing such techniques to address the current health challenges. The following sections discuss the related literature categorized based on the model output - height, weight, and medium of deployment.

\subsection{Height Prediction}
\textbf{Alberink et al. \cite{alberink2008obtaining}} pointed out that in the field of forensic practice, there is a recurring demand for height estimations of individuals observed in surveillance video footage captured by cameras. Multiple approaches exist for conducting such estimations and to gain insights into the disparities between actual and measured heights, validation measurements are taken from a group of test subjects. Based on this analysis, a method was proposed to determine confidence intervals for the height of individuals depicted in images, accounting for factors such as head and footwear. The aim was to provide a reliable framework for estimating the height of questioned individuals captured in surveillance images while considering both systematic and random sources of variation. Later, \textbf{Abdelkader et al. \cite{4813453}} employed an equation that predicts height based on explicitly labeled keypoint coordinates in the image. \textbf{Dey et al. \cite{10.1145/2660460.2660466}} assessed the height differences of individuals in every picture and generated a height disparity graph from a photo compilation to estimate height. Several of the earliest works estimated height and weight using metrics such as physique and bone length alongside face and body images. Then with the rise of deep learning, \textbf{Dantcheva et al. \cite{8546159}} first proposed a 50-layer ResNet architecture, achieving an 8.2 cm and 8.51 kg MAE for height and weight prediction, respectively, using only face images. \textbf{Gunel et al. \cite{gunel2019face}} later tried improving the architecture using face, body, and gender information for predicting height in unconstrained settings. In addition to these inputs, techniques involving depth information were developed, such as the work by \textbf{Fuken et al. \cite{9156541}}, where a four-stage architecture performs segmentation of the human body into explicit segments, predicts the height of the segments using three CNNs with an error of 0.9\% , and the research by \textbf{Lee et al. \cite{lee2020human}}, which devised a height estimation method using both color and depth information with the help of Mask R- CNN's, achieving a 2.2\% error rate. 



\subsection{Weight Prediction}
One of the initial works for weight estimation used anthropometric features as proposed by \textbf{Velardo et al. \cite{5634540}}. By employing multiple regression analysis, the authors aimed to establish a model that can effectively estimate weight using various anthropometric features. They relied on a comprehensive medical database to train the model, ensuring that it captures a wide range of anthropometric variations and provides accurate weight predictions. The weight assessor proposed by \textbf{Nguyen et al. \cite{nguyen2014seeing}} made use of the abundant information available in RGB-D images to improve estimation accuracy. The method takes into account visual color signals, depth information, and gender to estimate multiple weight-related dimensions. This integrated strategy offered an extensive framework for predicting mass from a single RGB-D image. Influenced by recent developments in health science research, \textbf{Jiang et al. \cite{8666768}} investigated the viability of analyzing body weight using 2D frontal view human body images with BMI as the metric for measuring body weight. The intention of the study was to examine this analysis at differing levels of difficulty by investigating three feasibility problems ranging from simple to complex. To facilitate the analysis of body weight from human body images, the researchers developed a system that involved computing five anthropometric features, which have been recommended as viable indices for determining body weight. A \textbf{visual-body-to-BMI dataset} has been acquired and systematically cleansed to support the research study. 

As mentioned previously, \textbf{Dantcheva et al. \cite{8546159}} investigated the viability of estimating measurements of height, weight, and BMI from single-shot photographs of the face. The authors proposed a regression method based on the 50-layer ResNet architecture to accomplish this goal. This method utilized the exclusive properties of facial images to precisely estimate the aforementioned characteristics. In addition, a new dataset containing 1026 subjects has been included in this study. In a recent study, \textbf{Jin et al. \cite{jin2022estimating}} noted that BMI is frequently employed as a measurement of weight and health conditions and that previous research in this field has focused primarily on using numerous 2D images, 3D images, or images of the face. However, these indicators are not always accessible and the authors proposed a dual-branch regression approach to estimate weight and BMI from a single 2D body image to circumvent this limitation. The researchers intend to improve the accuracy of BMI estimation from a single 2D body image by integrating information from the anthropometric feature computation branch and the deep learning-based feature extraction branch. In addition, few methods attempted to estimate both height and weight simultaneously, such as \textbf{Altinigne et al. \cite{9053363}}, who developed a deep learning method that employs the estimation of individual silhouette and skeleton joints as effective regularizers. 



\subsection{Malnutrition and IoT Solutions}

Many previous works have focused on developing a solution for malnutrition, such as the expert system by \textbf{Thapar et al. \cite{7906819}}, which analyses malnutrition using a Mamdani inference method with 13 different categorical input variables, but it is only recently that work has begun to make them accessible and deployable. One such IoT-based solution is \textbf{Child Growth Monitor \cite{Microsoft}}, an AI-based application that relies on the availability of infrared sensors in selected smartphones to capture 3D measurements of a child’s height, body volume, and weight ratio. However, even these techniques fell short of providing a complete solution involving height, weight estimation, all wrapped up in an application that could be used by anyone with a smartphone. Our work overcomes all of the aforementioned drawbacks while also improving weight estimation performance through the use of local 3D features, multimodal embedding fusion, and an edge device prototype for computation. Table \ref{Related Research Works} depicts an overview of all the discussed existing solutions.



\section{Methodology}

\begin{figure}[hbtp]
    \centering
    \includegraphics[width=\linewidth]{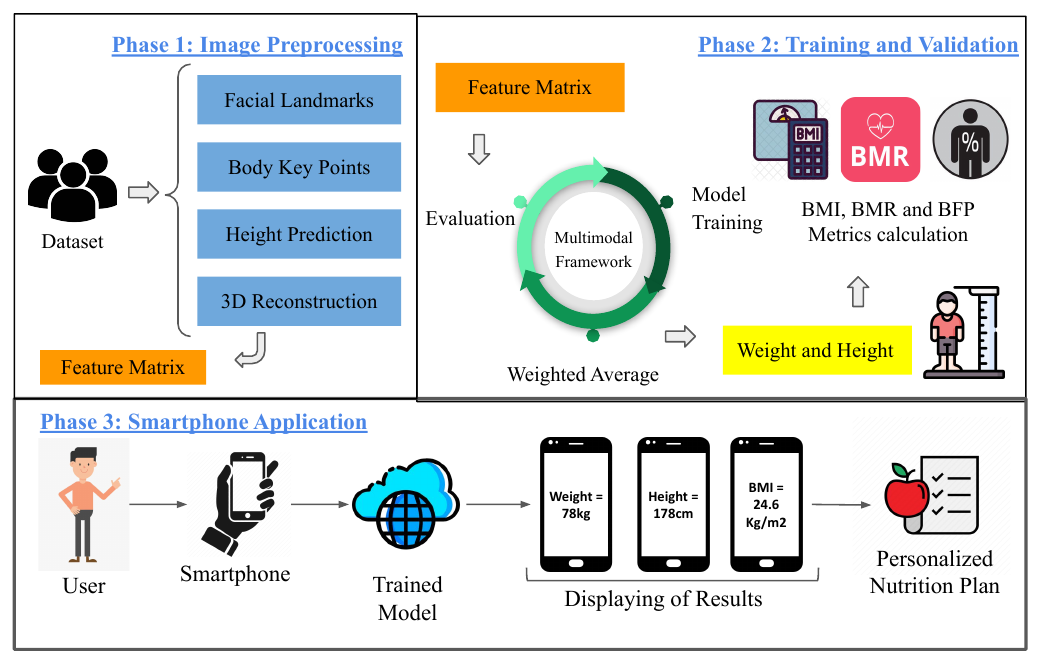}
    \caption{Proposed System Overview}
    \label{Methodology Overview}
\end{figure}

This section describes the proposed three-phase height and weight estimation workflow, as shown in Fig. \ref{Methodology Overview}. Phase 1 deals with image pre-processing and height estimation while phase 2 emphasizes feature extraction, multi-modal fusion, and regression. Subsequently, the final phase depicts the integration of the above system with an edge device application prototype in an IoT framework. 

\vspace{-3mm}

\subsection{Phase 1: Pre-processing and Height Prediction}

In this phase, we pre-process the input image of a person, reconstruct the 3D volumetric information and perform height prediction. The mentioned phase is divided into four sub-phases: Facial landmark detection and alignment, Body key points detection, 3D reconstruction and Height prediction. 

\subsubsection{Facial Landmarks Detection and Alignment}

\begin{figure}[hbtp]
    \centering
    \includegraphics[width=\linewidth]{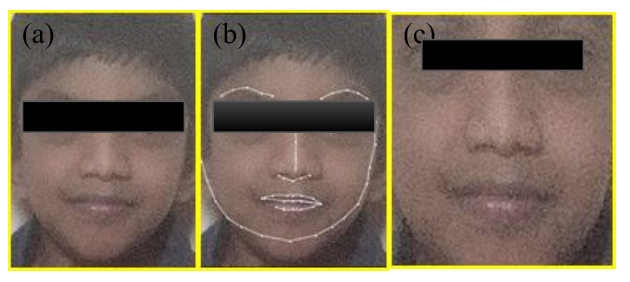}
    \caption{Our face verification and pre-processing pipeline: (a) Face Detection, (b) Facial Landmark Detection, (c) Face Alignment \& Cropping}
    \label{fig: face verification}
\end{figure}

To extract the face crop from full body image we perform face verification, cropping and subsequently alignment. The initial step of face detection determines the position of a face, by traversing through the points around the facial region to locate 68 landmarks. Subsequently, the faces are aligned and transformed such that facial landmarks (inner eyes and bottom lip) appear in approximately in same regions, preserving the collinearity, parallelism, and the ratio of distances between the points with Affine Transformation. Fig. \ref{fig: face verification} (a) visualizes an example of the localization of face from the input image, Fig. \ref{fig: face verification} (b) depicts the facial landmarks while Fig. \ref{fig: face verification} (c) illustrates the facial alignment and region cropping. After completing the facial alignment step, the subsequent stage in the preprocessing pipeline involves the detection of body key points.


\subsubsection{Body Keypoints Detection}

Considering the inherent unpredictability of real-world scenarios, it is imperative to establish the elimination of unwanted noise. Following the extraction of the human body region from varying backgrounds through the application of a U-Net trained for human segmentation, the subsequent stage involves the detection of human body landmarks within the input image. This process commences with the initial layers of the VGG-19 network extracting pertinent image features, which are then passed into two parallel branches of convolutional layers. The first branch predicts a group of 18 confidence maps, each representing a different portion of the human posture skeleton. The second branch predicts a group of 38 Part Affinity Fields (PAFs)  \cite{DBLP:journals/corr/abs-1812-08008}, which indicates the degree of affinity between parts. Let set S = $(S_{1}, S_{2},....,S_{J})$ denote the confidence maps for $j$ i.e., detected body parts. Then the individual confidence maps for each person $k$ can be formulated as $S^{*}_{j,k}$ at a location $p$ is denoted in the following Eq. \ref{OpenPose}, where $x_{j,k}$ be the ground truth position of body part $j$ for person $k$ in the image and $\sigma$ controls the spread of the peak. 

\begin{equation}
\label{OpenPose}
    S^{*}_{j,k}(p) = exp(- \frac{|p-x_{j,k}|^2_{2}}{\sigma^2})
\end{equation}

\subsubsection{3D Reconstruction}
The loss of 3D information during the process of capturing pictures poses a significant challenge in accurately inferring and extracting 3D characteristics from 2D visuals. To tackle the aforementioned challenge, we adopt a multi-level architecture PiFuHD \cite{DBLP:journals/corr/abs-2004-00452} which is trained end-to-end on high-resolution images. This model is profound in reconstructing 3D mesh, preserving intricate 3D details solely from a single human image.
The objective of the algorithm is to model a function, $f(X)$, such that for any given 3D position in continuous space $X = (X_{x}, X_{y}, X_{z}) \in R^3$, it predicts the occupancy value as shown in Eq. \ref{PFHD-1}.

\begin{equation}
\label{PFHD-1}
    f(X, I) = 
    \begin{cases}
       1,  \textit{if X is inside the mesh surface} \\
       0,  \textit{otherwise}
    \end{cases}
\end{equation}

For an orthogonal projected 2D point given by $\pi(X) = x = (X_{x}, X_{y})$, an image feature embedding is extracted by function $f$. Then the occupancy of the query 3D point X is estimated by Eq. \ref{PFHD-2} where Z = $X_{z}$ is the depth along the ray defined by the 2D projection $x$.

\begin{equation}
\label{PFHD-2}
    f(X, I) = g(\phi (X,I), Z)
\end{equation}

\begin{figure}[htbp]
    \centering
    \includegraphics[width=\linewidth]{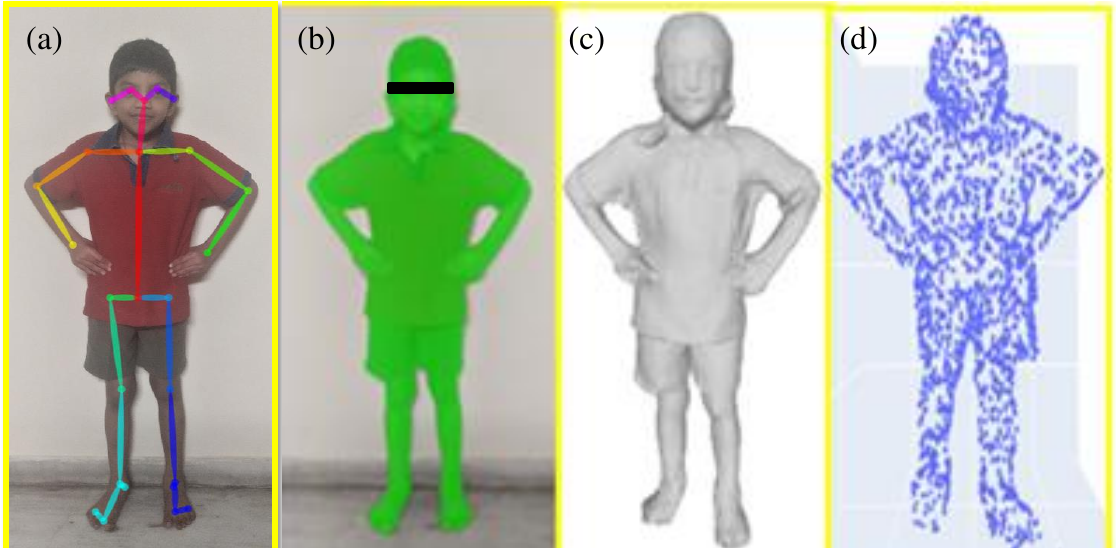}
    \vspace{0.01pt}
    \caption{Our body detection and prepossessing pipeline: (a) Body Keypoint estimation, (b) Masking, (c) 3D Human Mesh Reconstruction, (d) Conversion to 3D Point-Cloud}
    \label{Preprocessing}
\end{figure}

Finally, we employ mesh sampling to generate a point cloud representation of the mesh, which provides a straightforward yet efficient means of representing 3D data. The detected body key points are illustrated in Fig. \ref{Preprocessing} (a), Fig. \ref{Preprocessing} (b) depicts the result of masking the input image, Fig. \ref{Preprocessing} (c) shows the 3D Mesh Reconstruction and Fig. \ref{Preprocessing} (d) illustrates its conversion to 3D Point-cloud. 

\begin{figure*}[t]
    \centering
    \includegraphics[width=\textwidth]{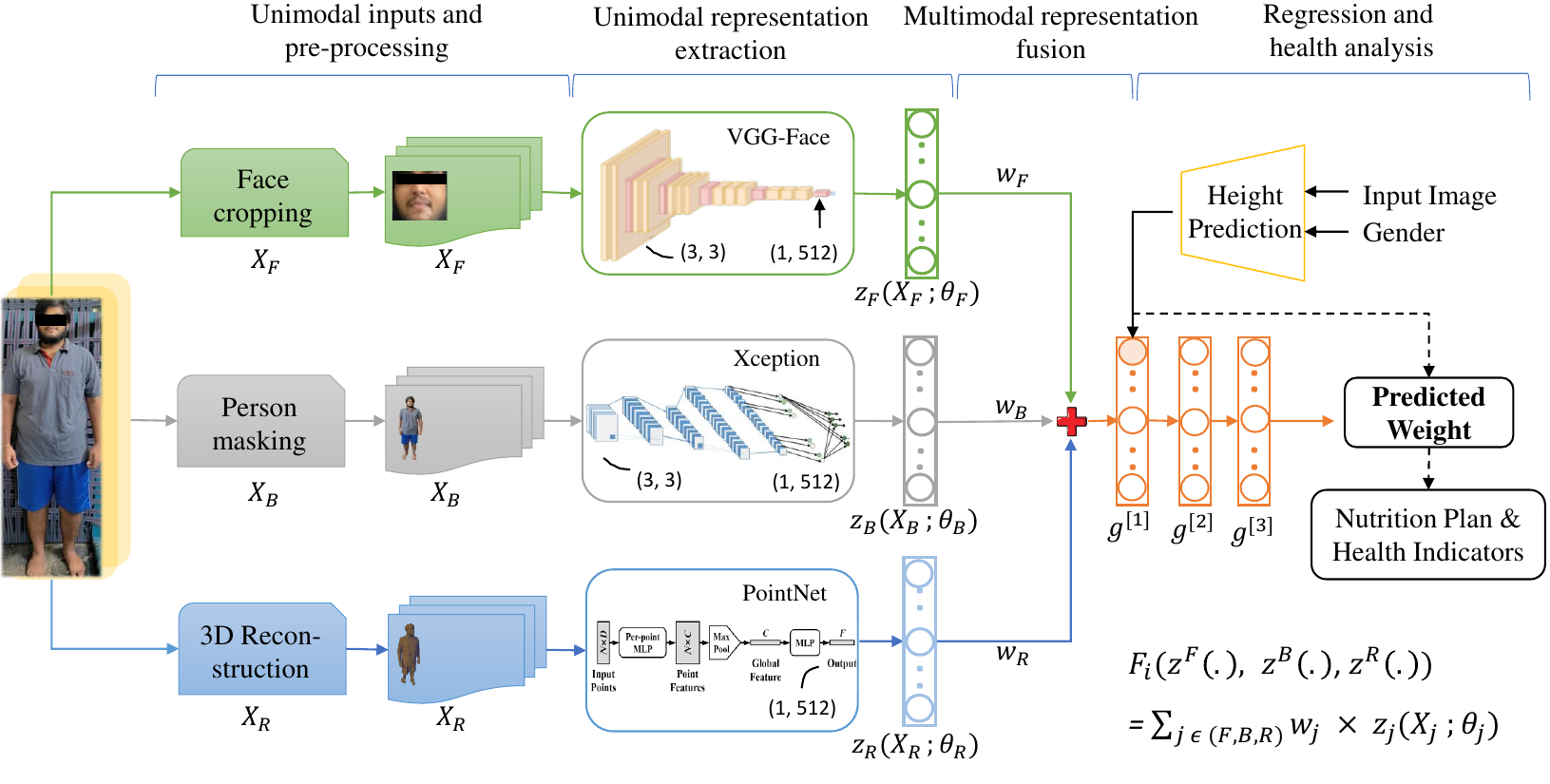}     \caption{Overview of our proposed multi-modal computational architecture. The feature fusion obtains the unimodal representations ${z_F}$, ${z_B}$, ${z_R}$ by passing the inputs ${X_F}$, ${X_B}$, ${X_R}$ into the sub-embedding networks parametrized by ${\theta_F}$, ${\theta_B}$, ${\theta_R}$ respectively. The representations are then weighed by learned weights ${w_F}$, ${w_B}$, ${w_R}$ and concatenated with gender and height information to predict the weight and subsequently calculate BMI, BMR, and BFP.}
    \label{Computational Architecture}
\end{figure*}

\subsubsection{Height Estimation}
The final step of this phase is height prediction, and taking previous work results into account, we decided to use a simple yet efficient computer vision technique that works best for input images that are parallel to the subject, similar to our dataset images. The simple pixel arithmetic method relies on the person's scale and camera orientation to calculate the person's height. To begin, we undistort the image to remove radial and tangential distortions and make the image independent of the device used to capture it. Then, we calculate the pixel per metric (ppm) attribute on the tight-crop masked image ($I_c$) from previous sub-phases using Eq. \ref{ppm}. This metric is then re-used throughout the process to predict the height of a new person ($I_{pred}$) given a static camera position by Eq. \ref{ppm-new}.


\begin{equation}
    \label{ppm}
    ppm = \frac{I_c.size[0]}{I_c \; height}
\end{equation}

\begin{equation}
    \label{ppm-new}
    height_{pred} = \frac{I_{pred}.size[0]}{ppm}
\end{equation}

\subsection{Phase 2: Unimodal representation and fusion}
The preprocessed data extracted from the previous phase is passed to this phase for feature extraction. This phase can be further divided into three sub-phases: 3D-feature extraction, 2D-feature extraction, multi-modal fusion and regression. The overview of the computational architecture is represented in Fig. \ref{Computational Architecture}

\subsubsection{3D feature extraction}
The point cloud obtained after the previous phase's pre-processing is used as an input to extract the 3D embedding representation. The 3D point classifiers are the best at classifying the point cloud based on the local granular shape and the overall global shape, making them the ideal feature extractors for our problem. As a result, we use the PointNet \cite{qi2017pointnet} classifier to compile depictions because of its capacity to deal with unordered input points by employing a symmetric function (max pooling) to learn a set of optimization functions/criteria that select informative areas in the point cloud and represent the explanation for their inheritance. The final fully connected layers of the network consolidate these optimally learned values into the global descriptor for the entire shape, resulting in the 512-dimensional feature vector. Since each point undergoes its own transformation, our input format makes it simple to implement unchanging or affine modifications.

\subsubsection{2D feature extraction}
The 3D embedding features have been computed in the previous step. Now we take a similar approach to calculate the 2D feature representation. First, the preprocessed face image is passed through a VGGFace architecture \cite{mehdipour2016comprehensive} without a head to extract a 512-dimension vector. Parallelly, we also pass the body image through an Xception architecture \cite{chollet2017xception} without a head, using it as a feature extractor to get a 512-dimension body representation. Here, the VGG-16 has 16 trainable convolutional layers followed by a max-pooling operation whereas Xception is a deep convolutional neural network architecture with Depthwise Separable Convolutions. Finally, we employ Transfer Learning techniques with these trained VGGFace and Xception model pre-trained weights to extract 2D facial and deep body features from preprocessed face and full-body images, respectively. This forms the basis for the subsequent step of multi-modal feature fusion and regression. 

\subsubsection{Multi-modal fusion and regression}
Now as all the unimodal features are extracted we fuse the different sub-embedding streams of 512- dimensional feature representations. 
    These representations comprises of two different modalities - point cloud ($z_R$) and image data ($z_F$, $z_B$) and hence cannot be fused with a simple concatenation. Instead, we use learnable weights ($w_F$, $w_B$, $w_R$) to weigh these features and add them all up to get a final 515-dimensional feature vector (line 2, Algorithm 1). This feature vector is then passed through two 512-units Multi Layer Perceptron (\emph{$g^{[0]}$}, \emph{$g^{[1]}$}), followed by 256 units MLP (\emph{$g^{[2]}$}) and finally through a single unit linear layer (\emph{$g^{[3]}$}) to predict the weight of the person (line 3-5, Algorithm 1). The final layer uses Ridge regression to penalize the layer to not overfit the distribution but to generalize to new plausible test data samples. Then we compute the person's Body Mass Index (BMI), followed by Body Metabolic Rate (BMR) using Mifflin-St Jeor Equation \cite{BMR} and Body Fat Percentage (BFP) using BMI for suggesting appropriate nutrition plan and malnutrition monitoring. In Algorithm 1 (lines 8-9), \emph{p} and \emph{m} are intercept constants that vary with gender, with values of 5 and 16.2 for men and 161 and 5.4 for women, respectively.

\begin{algorithm}
	\caption{: \emph{Multimodal fusion and Regression}}
	\begin{algorithmic}[1]
		\REQUIRE Input, $z_F$, $z_B$, $z_R$, \emph{gender}, $height_{pred}$
		\ENSURE $weight_{pred}$, \emph{BMI}, \emph{BMR}, \emph{BFP}\\[2pt]
		
		\STATE \textbf{$E({r_F}, {r_B}, {r_R})$} =  $\Sigma_{j \in (F,B,R)} \; w_j \times z_j(X_{j}; \; \theta_{j})$\\[2pt]
        \STATE $h^{[-1]}$(F, a, g) = concatenate(\emph{E, gender}, $height_{pred}$)\\[2pt]	
        \FOR{i in [0, 1, 2, 3]}
        \STATE $h^{[i]}$ = $g^{[i]}$($W^{[i]} \times h^{[i-1]}$ + $b^{[i]}$)
        \STATE $weight_{pred}$ = $h^{[i]}$ 
        \ENDFOR
        \STATE BMI = $\cfrac{weight_{pred}[kg]}{{height_{pred}}^2[m^2]}$ = $\cfrac{weight_{pred}[lb]\times 703}{{height_{pred}}^2[in^2]}$\\[4pt]	
        \STATE BMR = $10 \times weight_{pred}$ + $6.25 \times height_{pred}$ - $5 \times age + p$\\[2pt]	
        \STATE BFP = $1.2 \times BMI + 0.23 \times age - m$\\[2pt]	
	\end{algorithmic}
	\label{fusion-algo}
\end{algorithm}

\begin{table*}[h]
	\centering
	\caption{Statistical information of the Datasets}
	\label{datatable}
	\begin{tabular}{|c|c|m{4em}|c|c|m{4em}|m{4em}|m{7em}|m{6em}|m{4em}|m{4em}|m{6em}|}
		\hline
		\multicolumn{2}{|c}{\textbf{Participant Information}} & \multicolumn{2}{|c|}{\textbf{Gender}}  & \multicolumn{4}{c|}{\textbf{Height (in cm)}}  & \multicolumn{4}{c|}{\textbf{Weight (in kg)}}     \\
		\hline
		\centering Dataset &
		\centering Total &
		\centering Male & 
		\centering Female & 
		\centering Range & 
		\centering Mean & 
		\centering Standard Deviation & 
		\centering 95$\%$ Confidence Interval & 
		\centering Range & 
		\centering Mean & 
		\centering Standard Deviation & 
		\centering 95$\%$ Confidence Interval \tabularnewline
		\hline
		
		\centering Visual-body-to-BMI & 
		\centering 5900 & 
		\centering 3968 & 
		\centering 1932 & 
		\centering 213.36 - 147.32 & 
		\centering 175.54 & 
		\centering 9.89  & 
		\centering 176.99 - 174.09 & 
		\centering 254.01 - 44.90 & 
		\centering 95.05 & 
		\centering 27.12 & 
		\centering 100.9 - 89.1 \tabularnewline
		\hline
		
		\centering Locally Collected Data & 
		\centering 287 & 
		\centering 261 & 
		\centering 26 & 
		\centering 184-101 & 
		\centering 164.09 & 
		\centering 21.35  & 
		\centering 167.23 - 160.95 & 
		\centering 100 - 13 & 
		\centering 63.51 & 
		\centering 21.41 & 
		\centering 68.26 - 58.76 \tabularnewline
		\hline
	\end{tabular}
\end{table*}

\subsection{Phase 3: Android Application Prototype}

\begin{figure}[htbp]
    \centering
    \includegraphics[width=0.85\linewidth]{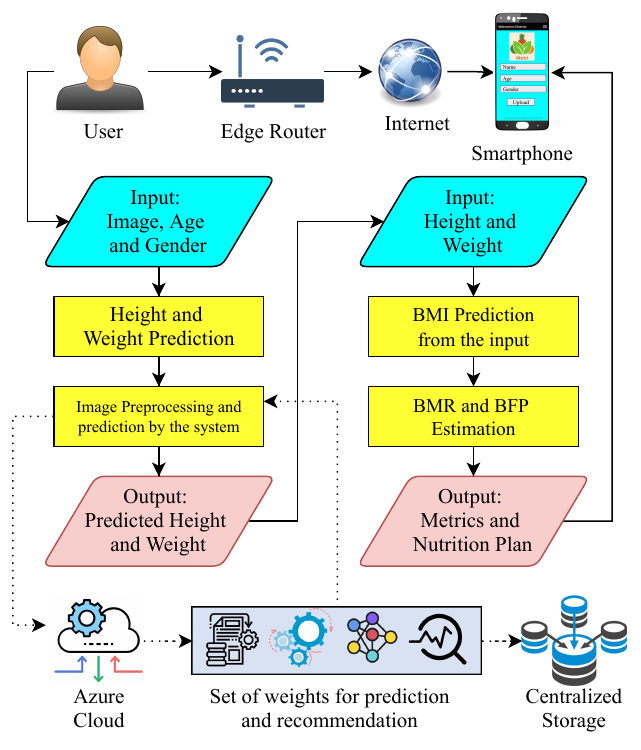}
    \caption{Workflow of the proposed IoT Application prototype}
    \label{Fig: Workflow}
\end{figure}

Following training the model, the model's learned weights are saved using Pytorch's.save() function and converted to an \emph{.pb} file using ONNX as the intermediate format \cite{9377615}. Then, we use TensorFlow Serving to deploy and serve the trained model as an \emph{.apk} file integrated with the created Android interface. The Android interface is intended to be simple and efficient for people from all walks of life and social strata. The workflow of the proposed system's GUI is depicted in Fig. \ref{Fig: Workflow}.

\section{Experimental Study} \label{exp-study}
This section describes the dataset used for training and testing our model, ablation studies, and experiments on the proposed multi-modal system under various scenarios. Following that, we will go over the performance of cloud-based IoT application as well as the computational platform used.

\begingroup
\setlength{\tabcolsep}{5pt} 
\begin{table*}[h]
	\centering
	\caption{Performance comparison of different architecture combinations for weight prediction}
	\label{arctable}
	\begin{tabular}{|m{4em}|m{6em}|m{4em}|m{4em}|m{4em}|m{4em}|m{4em}|m{4em}|m{4em}|m{4em}|m{4em}|}
		\hline
		\multicolumn{2}{|c}{\textbf{3D Features}} & \multicolumn{3}{|c|}{\textbf{PointNet}}  & \multicolumn{3}{c|}{\textbf{DG-CNN}}  & \multicolumn{3}{c|}{\textbf{GB-Net}} \\
		\hline

		\centering \textbf{Face FE} &
		\centering \textbf{Body FE} &
		\centering MAE &
		\centering RMSE &
		\centering {$R^2$} &
		\centering MAE &
		\centering RMSE &
		\centering {$R^2$} &
		\centering MAE &
		\centering RMSE & 
		\centering {$R^2$} \tabularnewline 
		\hline 

		\multirow{ 2}{*}{\textbf{VGGFace}} &
		\centering \textbf{Xception} &
		\centering 5.309 &
		\centering 7.438 &
		\centering 0.720 &
		\centering 7.763 &
		\centering 9.352 &
		\centering 0.572 & 
		\centering 6.421 & 
		\centering 8.396 & 
		\centering 0.639 \tabularnewline
        
		& \centering \textbf{ResNet152} &
		\centering 5.612 &
		\centering 7.635 &
		\centering 0.697 &
		\centering 7.894 &
		\centering 9.650 &
		\centering 0.560 & 
		\centering 6.989 & 
		\centering 8.903 & 
		\centering 0.596 \tabularnewline
		\hline
		
		\multirow{ 2}{*}{\textbf{FaceNet}} &
		\centering \textbf{Xception} &
		\centering 5.978 &
		\centering 7.998 &
		\centering 0.661 &
		\centering 8.363 &
		\centering 10.016 &
		\centering 0.559 & 
		\centering 6.640 & 
		\centering 8.511 & 
		\centering 0.615 \tabularnewline
        
		& \centering \textbf{ResNet152} &
		\centering 6.112 &
		\centering 8.131 &
		\centering 0.651 &
		\centering 8.606 &
		\centering 10.400 &
		\centering 0.548 & 
		\centering 7.200 & 
		\centering 9.155 & 
		\centering 0.587 \tabularnewline
		\hline
		
	\end{tabular}
\end{table*}
\endgroup

\subsection{Datasets Used}

In our work, we majorly used two datasets - visual-body-to-BMI dataset \cite{8666768}, locally collected dataset. As mentioned earlier, the visual-body-to-BMI dataset consists of 47574 images of 16483 people scraped and downloaded from the progresspics subreddit website. These images are then annotated and filtered resulting in a total of 5900 images, with two images for each of the 2950 subjects. The 2950 subjects comprises of 966 females and 1984 males, as well as the corresponding gender and weight labels. On the other hand, we locally collected a dataset of 30 people in 9 - 10 frontal poses, as well as height and device information. Table \ref{datatable} highlights the statistical information about these two datasets. These two datasets are then combined to jointly train the model but is bench-marked only on the visual-body-to-BMI to enable comparison with the previous works. Meanwhile, we have held-out a sample size of 30 from the locally collected dataset, with each image being one of the 30 subjects in a randomly sampled pose, for experiments across devices and lighting conditions in Section \ref{exp-study} D, \ref{exp-study} G respectively.

\vspace{-2mm}

\subsection{Steps followed to  capture input images}
The following steps are followed while capturing a full-body image of a person to estimate height and weight:
\begin{itemize}
    \item An RGB image of a frontal pose of person standing at a distance of 1.5 meters from the camera lens placed 1 meter from the ground is captured under sufficient lighting conditions as depicted in Fig. \ref{SR Results} (a).
    \item The smartphone lens was parallel to the person, i.e.,  90-degree angle w.r.t the person, and perpendicular to the ground, to accurately calculate the per-pixel metric for height estimation.
    \item The captured image is further masked \& pre-processed to remove the redundant background thereby extracting the facial, body, and 3D representations under pre-processing \& feature extraction pipelines. 
\end{itemize}

\subsection{Performance of multiple model architecture combinations}

To come up with the current architecture, we systemically explored the combinations of various facial feature extractors like VGGFace and FaceNet \cite{7298682}, body feature extractors like Xception and ResNet-152 in combination with 3D feature extractors like PointNet \cite{pointnet}, DG-CNN and GB-Net as summarized in Table \ref{arctable}. The best architecture observed is a combination of Xception, VGG-Face and PointNet for the body, face, and 3D feature extraction, achieving a MAE weight of 5.3 kg. We also noticed that VGG-Face outperforms FaceNet in general, while Xception outperforms ResNet-512. PointNet, on the other hand, outranks its corresponding point-cloud classifiers with its ability to extract rich 3D representations.


\subsection{Effect of Lighting Conditions on height and weight prediction and Device Comparison}

\begin{figure}[htbp]
    \centering
    \includegraphics[width = \linewidth]{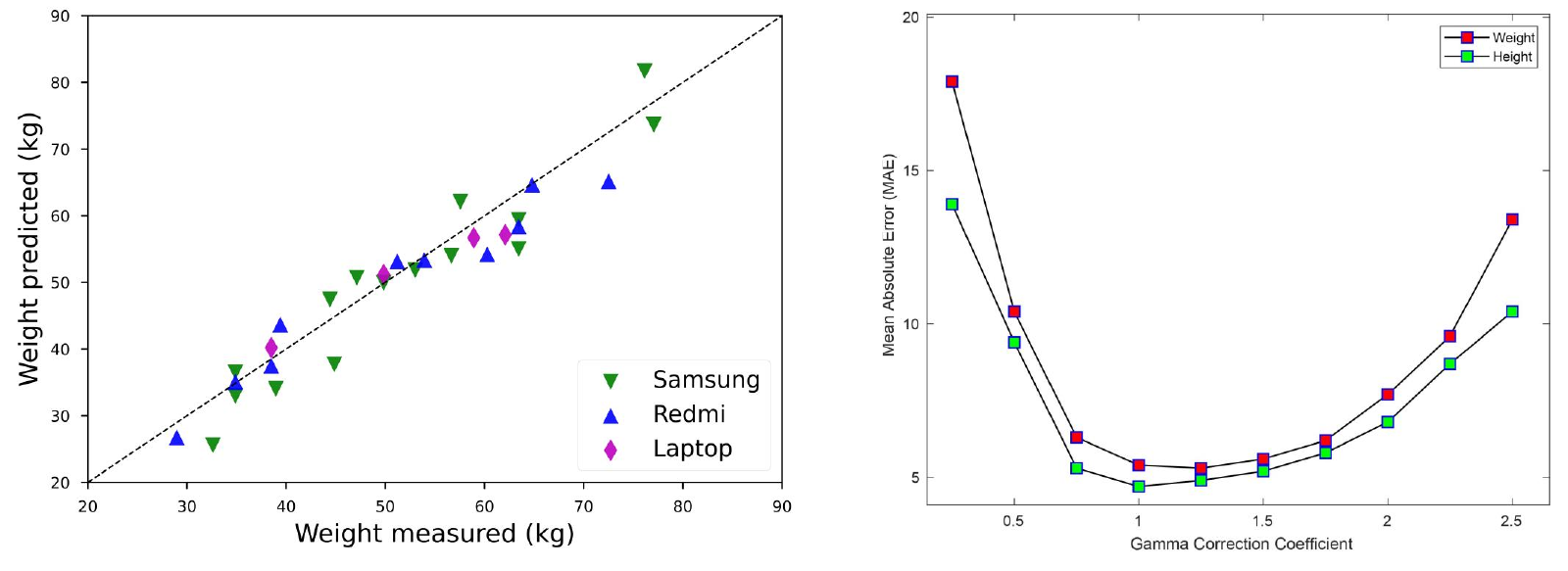}
    \caption{(a) Performance under simulated lighting conditions. Our system remains robust to wide range of illumination but it is preferable to have sufficient lighting to decrease the error. (b) Performance of our system across various devices}
    \label{DCL}
\end{figure}

The collected dataset contains images in unconstrained lighting conditions and is a perfect representation of real-world lighting conditions. To illustrate this and test the model performance further, we have artificially simulated the image brightness using gamma correction. The model performs best in $\gamma$ range of 1.0 to 1.25 as shown in Fig. \ref{DCL} (b). 
We can also deduce that the MAE decreases when $\gamma$ is in the range of 0 - 1.25, attains its minimum MAE at $\gamma$ = 1.0, and increases as $\gamma$ increases. The above variation in extreme cases can be attributed primarily to the poor performance of 3D reconstruction in extreme lighting conditions, where reconstruction quality decreases considerably when image global lighting drastically increases or decreases, despite performing well for a wide range of natural illumination.
For performance comparison on different devices we used a hold-out set which contains images collected from a variety of devices, including laptops and multiple smartphone brands. Figure \ref{DCL} (a) shows the predicted weight versus the actual weight, demonstrating that our model's performance is robust and coherent across all types of devices. 

\vspace{-6mm}

\subsection{Importance of Multiple Features}
\begin{figure}[htbp]
    \centering
    \includegraphics[width = \linewidth]{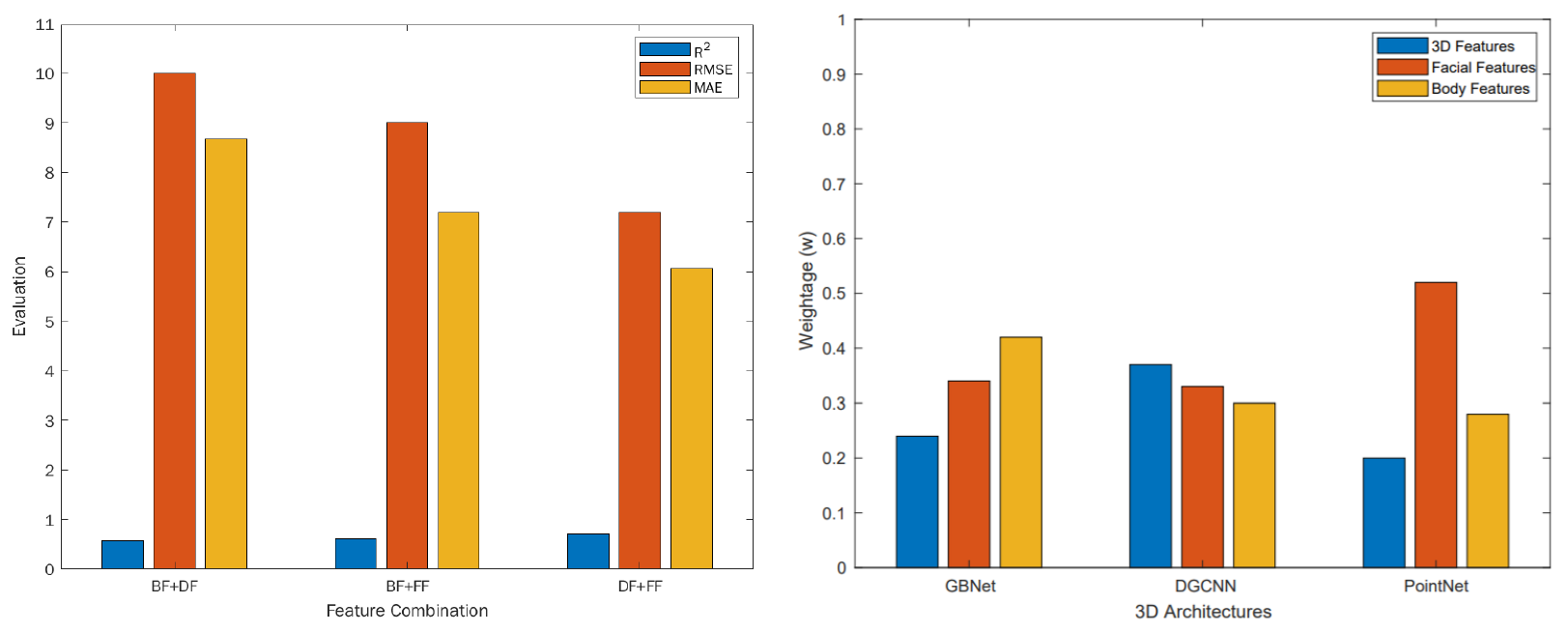}
    \caption{(a) Performance comparison of pairwise feature combinations in weight estimation, (b) Feature importance across different 3D network architectures.}
    \label{FI}
\end{figure}

Our best-performing model works on weighing and averaging the multiple input feature embeddings. The embeddings are weighed such that each embedding is assigned a weight between 0 and 1, and their sum equals 1. It enables us to interpret the relative importance of these different embeddings across multiple architectures in predicting the weight. We have observed that these weights vary significantly when the 3D feature extractor architecture is changed, while the best extractors for both facial and body features are kept constant. From Fig. \ref{FI} (b), we can also infer that PointNet allows the model to have a balanced weight distribution with lower error as compared to the others. Overall, though the 3D features have relatively low importance, they perform slightly better in extreme use-cases such as obese and under-nourished conditions than only-image-based techniques. 

Furthermore, in order to find the best pair-wise feature combination, we systematically tested combinations of various types of features, as shown in Fig. \ref{FI} (a).
The abbreviations BF, DF \& FF in Fig. \ref{FI} (a) represent the body features (BF), 3D features (DF) \& facial features (FF) respectively. This experiment is carried out by including the best architectures (PointNet+Xception+VGG-Face) for the respective features.
The best pair-wise combination of DF+FF, with low MAE and high correlation, demonstrates the importance of 3D and facial features. Despite the lack of BFs, the combined effect of DFs and FFs can still yield a reasonable result. The presence of scale free images in the dataset, which does not produce meaningful human structure anthropometric representations, can be attributed in large part to the lack of BFs importance.
To summarise the previous experiments findings, we can conclude that facial features are the most important predictor across all possible architecture combinations, followed mostly by 3D features and body features.

\begin{figure}[htbp]
    \centering
    \includegraphics[width=\linewidth]{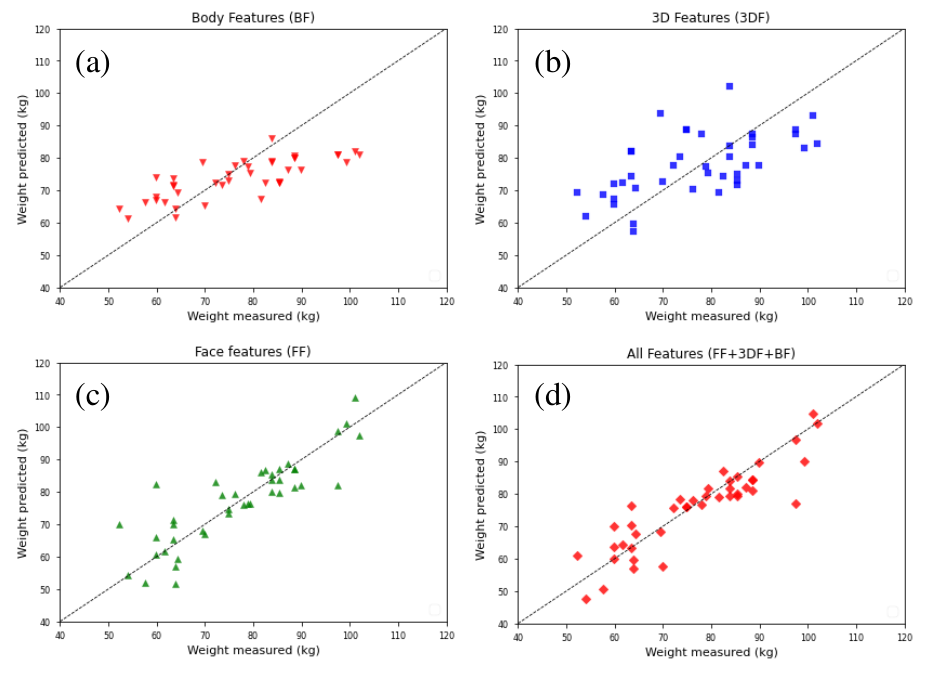}
    \caption{Correlation between predicted and measured body weight of 42 randomly selected test-set samples using only: (a) Body Features (BFs), (b) 3D Features (DFs), (c) Facial Features (FFs) and, (d) Combination of all features (FFs+DFs+BFs).}
    \label{BMI Correlation Plots}
\end{figure}

In addition, we illustrated the correlation plots of the individual features including BFs, DFs \& FFs along with their combination on our best found architecture in Fig. \ref{BMI Correlation Plots} (a), Fig. \ref{BMI Correlation Plots} (b), Fig. \ref{BMI Correlation Plots} (c) and Fig. \ref{BMI Correlation Plots} (d).

\vspace{-3mm}

\subsection{Android Application and Deployment}
\begin{figure}[htbp]
    \centering
    \includegraphics[width=\linewidth]{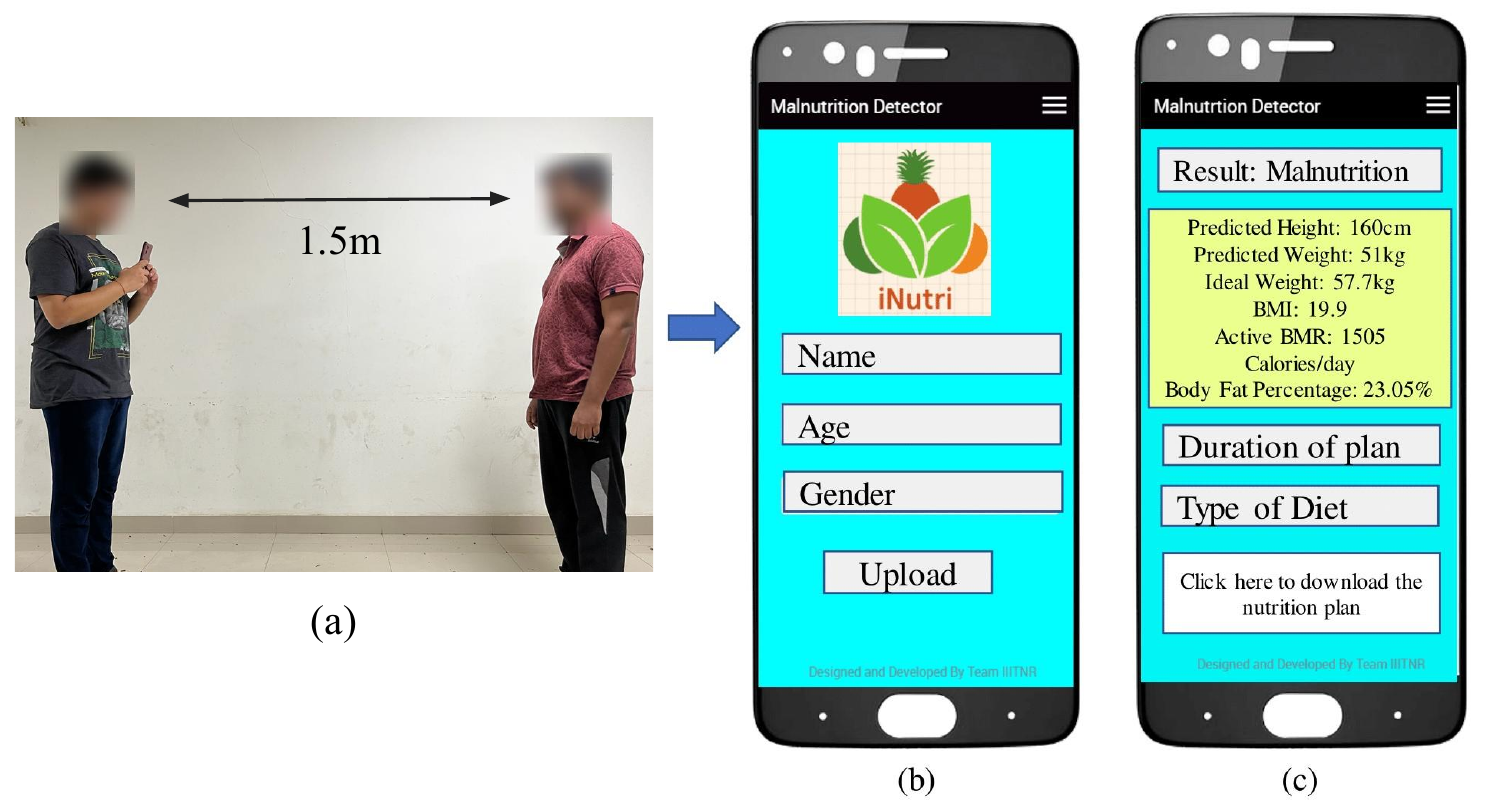}
    \caption{(a) Image acquiring technique, (b) Uploading the picture in addition to the associated metadata, (c) Illustrating the calculated outcomes and choosing a diet approach.}
    \label{SR Results}
\end{figure}

To interface with the proposed model, we designed a versatile and user-friendly android application.
The first page of the Graphical User Interface (GUI) consists of three sets of inputs: Age, Gender, and Image of the subject as shown in Fig. \ref{SR Results} (b). Next, the inputs are provided and the model present in the cloud computes the different output metrics. These computed metrics include the height, weight, BMI, Ideal weight, active BMR and the BFP of the person. To achieve the ideal weight, the user is then asked to select the type of diet and the number of weeks they are willing to dedicate to the program to attain the desired weight as shown in Fig \ref{SR Results} (c). Once this computation is performed, the results are displayed and the customized nutrition plan is made ready to download.

\subsection{Performance Observation on real-time data}
We further extensively tested our system for malnutrition classification on held-out locally collected real-time data from 30 people in various frontal poses. Based on their true BMI values, 20 of these 30 participants are healthy, while the remaining 10 are considered malnourished. The model's corresponding confusion matrix on this withheld dataset is as shown in Table \ref{ROCval}. As depicted, the model achieves an accuracy of 86.67\%, as well as precision, recall, and F1 score of 80 \%, 80 \%, and 80 \%, respectively.


\begin{table}[h]
	\caption{Confusion matrix of Malnutrition classification on Testset}
	\begin{center}
		\begin{tabular}{|c|c|c|c|c|c|c|}
			\hline
			\multirow{2}{*}{\textbf{Predicted Condition}} & 
			\multicolumn{2}{c|}{\textbf{Actual Condition}} & 
			\multirow{2}{*}{\textbf{Accuracy}} &
			\multirow{2}{*}{\textbf{Precision}} & 
			\multirow{2}{*}{\textbf{Recall}}   	\\
			\cline{2-3}
			& Healthy & Malnutritious & & & \\
			\hline
			Healthy & 18 & 2 &\multirow{2}{*}{86.67 \%} &\multirow{2}{*}{80.00 \%} &\multirow{2}{*}{80.00 \%} \\
			\cline{1-3}
			Malnutritious & 2 & 8 & & & \\
			\hline
		\end{tabular}
	\end{center}
	\label{ROCval}
\end{table}

\section{Key findings and Comparative Analysis}
In the proposed solution, several architectures with a combination of different fusion techniques for weight estimation and a pixel per metric approach for height estimation have been extensively tested. These findings are then used to calculate and infer pertinent health indicators from a single image, ultimately determining if the person is malnourished. The following are the key findings and comparative analysis of the proposed solution:


\subsection{Key Findings}
\textbf{Fine-scale 3D Representation}: 
Our research presents a pioneering application of PiFuHD for reconstruction, employing highly precise and detailed local 3D representation. This approach allows for a fine-grained level of detail in the reconstructed output. Furthermore, we utilize state-of-the-art 3D classification networks in our work by removing the last layers to extract a 512-dimensional vector as the 3D feature embedding. This technique enables us to capture and represent essential information from the input data.

\textbf{Multi-Modal Learning Paradigm}:
Many existing solutions in the field often rely on a single modality or feature representation, such as facial or manually crafted anthropometric information, or statistical measures, as highlighted in previous studies \cite{8546159} \cite{8666768} \cite{WEN2013392} \cite{JIANG2019183} \cite{9428234}. However, in our research, we adopted a holistic feature representation approach and conducted a systematic exploration to ascertain the significance of various features. This was achieved through extensive experimentation and in-depth analysis. Our solution stands out by achieving state-of-the-art results in weight estimation. Notably, we achieved the lowest mean absolute error (MAE) of 5.3 kg, surpassing previous works. This achievement was made possible by employing learnable weighing parameters in fusion, which enhances the accuracy of our weight estimation model. Through our research, we provide a comprehensive and advanced approach to weight estimation, considering multiple features and their interplay.

\textbf{Edge Device Deployment:}
A notable observation in the existing literature is the absence of deployed solutions or a reliance on sensor infrastructure for collecting user health data for monitoring, as highlighted in previous studies \cite{7906819} \cite{9011599} \cite{8449987}. In contrast, our research introduces a novel solution through the development of a smart application prototype. This prototype enables the estimation of health parameters such as height and weight and predicts the risk of malnutrition using a single full-body image. Importantly, this solution proves particularly valuable in remote locations with limited or no access to health facilities. One significant advantage of our approach is the use of an edge device prototype that operates independently, eliminating the need for additional equipment. This self-sufficiency empowers the prototype to estimate nutritional status accurately, providing crucial health insights even in resource-constrained environments.

\subsection{Comparative Analysis}

The proposed methodology showcased remarkable performance, achieving an impressive mean absolute error (MAE) score of 5.3 kg in weight estimation and 4.7 cm in height estimation. A comprehensive evaluation of error rates in height and weight prediction revealed that our approach outperformed previous works \cite{8546159} \cite{8666768} \cite{9053363} \cite{Microsoft}, as highlighted in Table \ref{MAE Comparison}. Importantly, our designed multi-modal system operates autonomously, eliminating the need for human intervention during crucial stages such as detecting body and facial landmarks, masking, cropping, and alignment. This autonomy enhances the efficiency and reliability of the system, setting it apart from non-autonomous and non-invasive techniques.

\begin{table}[htbp]
\caption{Mean Absolute Error (MAE) comparison with existing techniques}
\begin{center}
\resizebox{\columnwidth}{!}{%
\begin{tabular}{|c|c|c|c}
\hline
\textbf{Method}  & \textbf{Height MAE} & \textbf{Weight MAE}  \\ \hline
Altinigne et al.\cite{9053363}& 
$\pm$ 6.13 cm & 
$\pm$ 9.80 kg \\ 
\hline

Dantcheva et al.\cite{8546159}&
$\pm$ 8.2 cm & 
$\pm$ 8.51 kg \\
\hline

Jiang et al. \cite{8666768}&
-&
-& \\ 
\hline

Child Growth Monitor \cite{Microsoft}&
-&
-& \\
\hline

\textbf{Ours}& 
\textbf{$\pm$ 4.7 cm}& 
\textbf{$\pm$ 5.3 kg} \\ 
\hline

\end{tabular}
}
\end{center}
\label{MAE Comparison}
\end{table}

\section{Conclusion and Future Works}
This research presents a novel approach for predicting height and weight and inferring other health indicators, such as BMI, BMR, and BFP, from a single-shot full-body image. The methodology employs a holistic feature representation within a multi-modal learning paradigm. The proposed solution undergoes meticulous validation and testing using real-world images, including the simulation of various lighting conditions. The study also systematically examines the significance of 2D and 3D features. To further enhance the performance of weight and height prediction, future investigations can explore more rigorous methods for training and converging the multi-modal architecture. Additionally, efforts can be made to improve the extraction of FFs (Feature Fusion), DFs (Depth Fusion), and BFs (Body Fusion) embeddings. Exploring sub-embedding representation fusion methods and designing approaches to predict height without scale information or constraints could also contribute to improved prediction accuracy. Furthermore, future app development endeavors can focus on fostering communities and addressing security concerns related to the Machine Learning model and databases. These aspects will contribute to a more comprehensive and impactful implementation of the solution.

\section{Acknowledgements}
The authors would like to thank Dr. Min Jiang for providing access to the Visual-body-to-BMI dataset for our research. The authors are also grateful to the 30 volunteers for the contribution of the required images in creating a local dataset.

\bibliographystyle{ieeetr}
\bibliography{sHEMO_Biblography}
\vspace{-2cm}

\end{document}